\documentclass[conference]{IEEEtran}
\IEEEoverridecommandlockouts
\usepackage{cite}
\usepackage{makecell}
\usepackage{amsmath,amssymb,amsfonts}
\usepackage{algorithmic}
\usepackage{graphicx}
\usepackage{textcomp}
\usepackage{xcolor}
\usepackage{multirow}
\usepackage{array}
\usepackage{booktabs}
\usepackage{hyperref}
\usepackage{float}
\usepackage{caption}
\usepackage{subcaption}
\usepackage{placeins}
\usepackage{pdfpages}
\def\BibTeX{{\rm B\kern-.05em{\sc i\kern-.025em b}\kern-.08em
    T\kern-.1667em\lower.7ex\hbox{E}\kern-.125emX}}

\begin{document}

\title{GateAttentionPose: Enhancing Pose Estimation with Agent Attention and Improved Gated Convolutions}

\author{
\IEEEauthorblockN{1\textsuperscript{st} Liang Feng\thanks{*Corresponding author}}
\IEEEauthorblockA{\textit{Shenzhen University} \\
Shenzhen, China \\
q975638987@gmail.com}
\and
\IEEEauthorblockN{2\textsuperscript{nd} Zhixuan Shen\thanks{†These authors contributed equally as second authors}}
\IEEEauthorblockA{\textit{Shenzhen University} \\
Shenzhen, China \\
2310485003@email.szu.edu.cn}
\and
\IEEEauthorblockN{2\textsuperscript{nd} Lihua Wen\textsuperscript{†}}
\IEEEauthorblockA{\textit{Shenzhen University} \\
Shenzhen, China \\
2210485002@email.szu.edu.cn}
\and
\IEEEauthorblockN{2\textsuperscript{nd} Shiyao Li\textsuperscript{†}}
\IEEEauthorblockA{\textit{Shenzhen University} \\
Shenzhen, China \\
2310485006@email.szu.edu.cn}
\and
\IEEEauthorblockN{Ming Xu\textsuperscript{*}}
\IEEEauthorblockA{\textit{Shenzhen University} \\
Shenzhen, China \\
xuming@szu.edu.cn}
}

\maketitle

\begin{abstract}
This paper introduces GateAttentionPose, an innovative approach that enhances the UniRepLKNet architecture for pose estimation tasks. We present two key contributions: the Agent Attention module and the Gate-Enhanced Feedforward Block (GEFB). The Agent Attention module replaces large kernel convolutions, significantly improving computational efficiency while preserving global context modeling. The GEFB augments feature extraction and processing capabilities, particularly in complex scenes. Extensive evaluations on COCO and MPII datasets demonstrate that GateAttentionPose outperforms existing state-of-the-art methods, including the original UniRepLKNet, achieving superior or comparable results with improved efficiency. Our approach offers a robust solution for pose estimation across diverse applications, including autonomous driving, human motion capture, and virtual reality.
\end{abstract}


\section{Introduction} 
Human pose estimation, a key challenge in computer vision, has broad applications in autonomous driving, motion capture, and virtual reality. Despite recent advancements significantly improving accuracy, balancing high precision and computational efficiency remains a critical challenge.

\begin{figure}[htbp]
    \begin{minipage}[b]{0.5\textwidth}
        \centering
        \includegraphics[width=\linewidth]{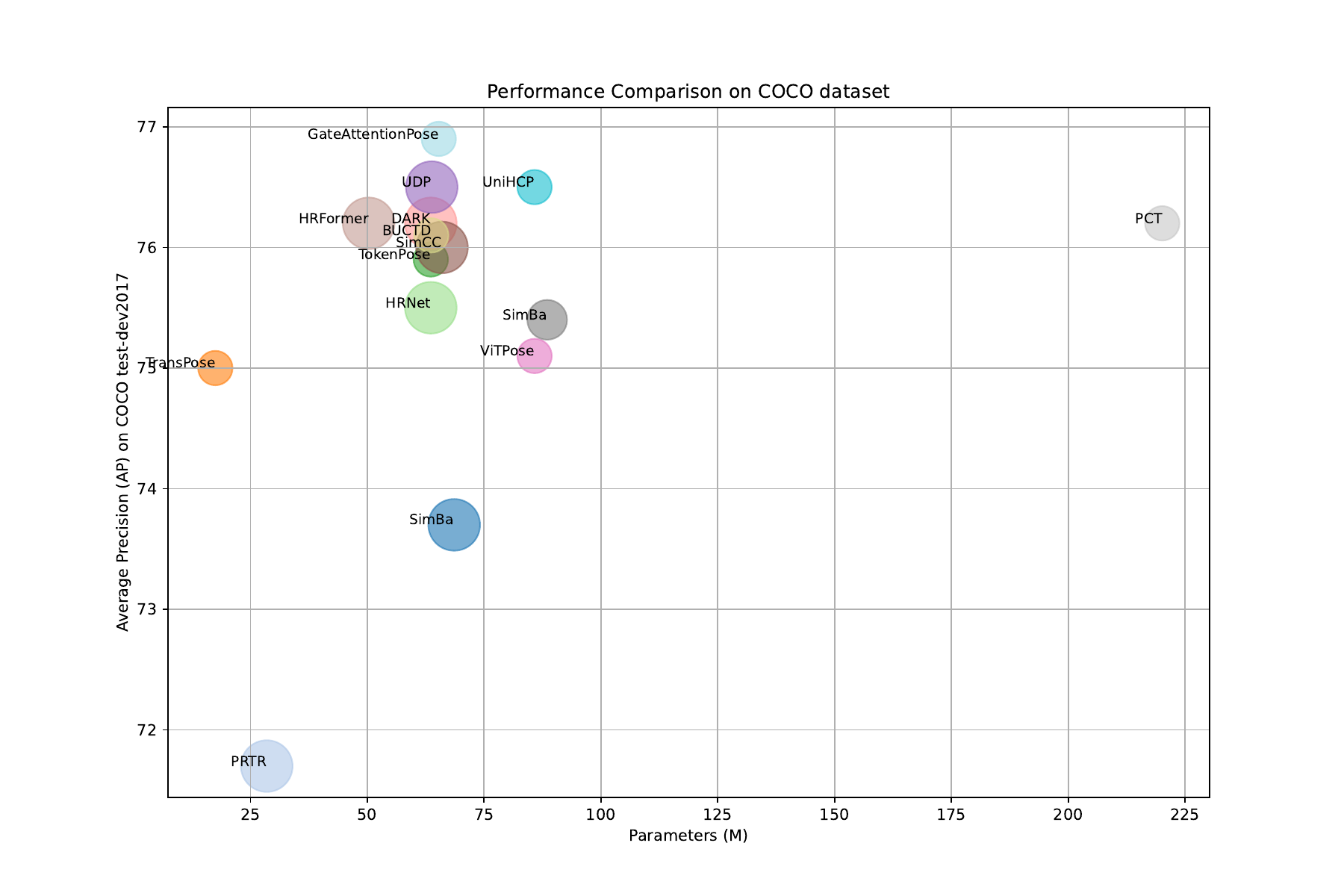}
        \caption{The comparison of GateAttentionPose and advanced methods on the COCO test-dev2017 set regarding model size and precision. The size of each bubble represents the input size of the model.}
        \label{fig:performance_comparison}
    \end{minipage}
\end{figure}

Traditional methods often struggle to capture complex joint relationships, leading to implausible predictions. In contrast, the human visual system excels at inferring holistic poses using contextual cues, highlighting the crucial role of context in achieving robust and accurate estimation.

We propose an innovative framework that enhances both accuracy and computational efficiency. Our approach includes: \begin{enumerate} \item Introduction of the GEFB module, improving feedforward operations while reducing parameters. \item Refinement of large kernel convolutions for superior feature extraction. \item Integration of Agent Attention to boost computational efficiency. \item Extensive evaluation on COCO and MPII datasets, demonstrating competitive performance with lower model complexity. \end{enumerate}

Our research provides a robust and efficient solution for pose estimation, striking an optimal balance between accuracy and resource utilization. This advancement has the potential to propel both theoretical understanding and practical applications in the field.

\section{Related Work} 
\subsection{Pose Estimation Approaches}

Pose estimation methods are categorized into top-down and bottom-up approaches. Top-down methods \cite{fang2017rmpe,Chen2020.12.04.405159,xu2022vitpose} detect individuals before estimating poses, while bottom-up methods \cite{cao2017realtime,cheng2020higherhrnet} detect all body parts and group them into instances. Recent transformer-based approaches \cite{yang2021transpose,xu2022vitpose} have shown promising results.

The COCO \cite{lin2014microsoft} and MPII \cite{andriluka20142d} datasets serve as widely adopted benchmarks, offering diverse scenarios for evaluating pose estimation methods across real-world challenges.

\subsection{Challenges in Crowded Scene Pose Estimation}

Pose estimation in crowded scenes faces challenges due to occlusions and dense arrangements. Studies like MIPNet \cite{khirodkar2021multi} and PETR \cite{shi2022end} have addressed these issues using hybrid approaches and transformer-based decoders.

\subsection{Innovations in Attention Mechanisms and Convolutional Techniques}

Our work adapts Agent Attention \cite{han2023agent} to replace large kernel convolutions, enhancing computational efficiency while preserving global context modeling. We introduce the Gate-Enhanced Feedforward Block (GEFB), building upon gated convolution techniques \cite{dauphin2017language} to improve feature extraction efficiency. These innovations address limitations in handling complex scenes and occlusions, demonstrating significant improvements in accuracy and efficiency on COCO and MPII datasets.

\section{Methodology}

\subsection{Overall Architecture}

\begin{figure*}[htbp]
    \centering
    \begin{subfigure}[b]{0.8\textwidth}
        \centering
        \includegraphics[width=0.8\textwidth]{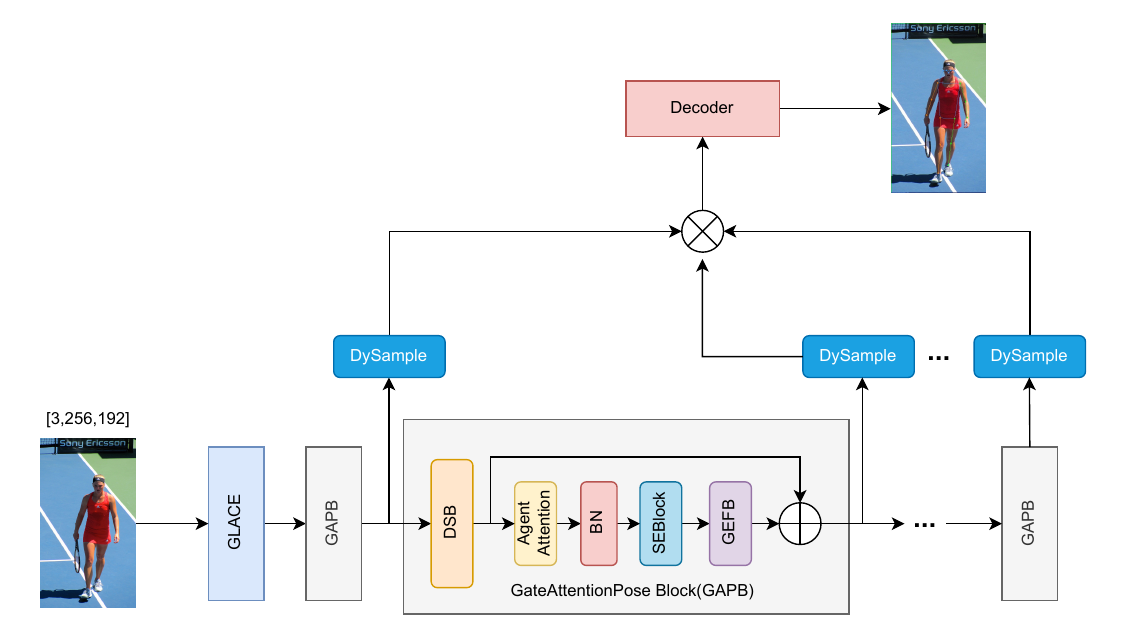}
        \label{fig:model_architecture}
    \end{subfigure}
    
    \begin{subfigure}[b]{0.3\textwidth} 
        \centering
        \includegraphics[width=\textwidth]{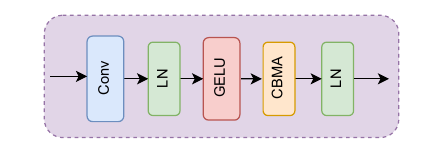}
        \caption{Downsample Block}
        \label{fig:dsb}
    \end{subfigure}
    \hfill
    \begin{subfigure}[b]{0.3\textwidth} 
        \centering
        \includegraphics[width=\textwidth]{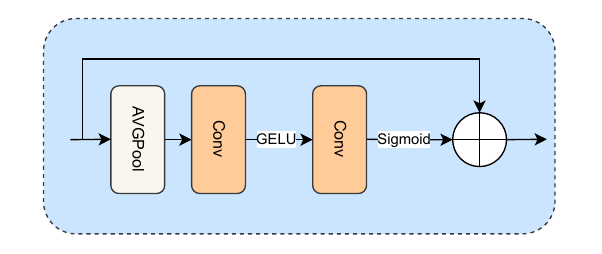}
        \caption{SENet Block}
        \label{fig:seblock}
    \end{subfigure}
    \hfill
    \begin{subfigure}[b]{0.3\textwidth} 
        \centering
        \includegraphics[width=\textwidth]{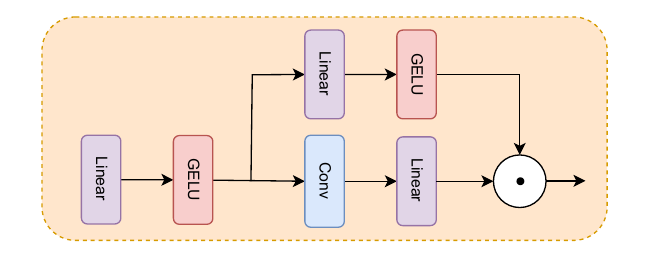}
        \caption{Gate-Enhanced Feedforward Block}
        \label{fig:gefb}
    \end{subfigure}
    
    \caption{The overall network architecture of our GateAttentionPose, as well as the (a) Downsample Block (DSB), the (b) SENet Block (SEBlock), and the (c) Gate-Enhanced Feedforward Block (GEFB).}
    \label{fig:all_architectures}
\end{figure*}

The architecture of our proposed model, illustrated in Figure \ref{fig:all_architectures}, comprises several key components. Initially, the input image, dimensioned [3, 256, 192], undergoes processing through the GLACE module \cite{wang2024glace}. This module embeds the image into a feature map of size [96, 64, 48], which subsequently serves as input to our enhanced backbone. The backbone, specifically engineered to efficiently handle complex scenes, further processes this embedded representation.

\subsection{GLACE Module Optimization}

We adapt the GLACE module \cite{wang2024glace} for our task, fine-tuning its parameters to transform 2D input images from $[3, 256, 192]$ to feature maps of $[96, 64, 48]$. The input image undergoes a series of convolutional layers, carefully adjusted to balance spatial resolution reduction and critical feature preservation. This optimized pipeline embeds the image into a feature map $F \in \mathbb{R}^{96 \times 64 \times 48}$.

Key modifications include:
\begin{itemize}
    \item \textbf{Convolutional Kernel Refinement}: Optimizing dimensions and quantity of kernels to align with input data properties and task requirements.
    \item \textbf{Activation Function Selection}: Adopting the most effective nonlinear activation function through extensive experimentation.
    \item \textbf{Adaptive Pooling Strategy}: Reconfiguring pooling layers to balance feature preservation and spatial dimensionality reduction.
\end{itemize}

These targeted adjustments enable the GLACE module to mitigate input data redundancy and extract more salient, task-relevant features, enhancing subsequent pose estimation processes.

\subsection{Advanced Feature Extraction Backbone}

Our model's backbone architecture integrates synergistic components that optimize feature extraction and computational efficiency, elevating performance across diverse scenarios.

\begin{itemize}
    \item \textbf{Downsample Block}: Halves spatial dimensions while doubling channel depth, incorporating a Convolutional Block Attention Module (CBAM) \cite{woo2018cbam} to refine feature representation by focusing on salient regions.
    \item \textbf{Agent Attention}: Replaces large kernel convolutions with an Agent Attention module \cite{han2023agent}, enhancing computational efficiency while maintaining global context modeling and capturing long-range dependencies.
    \item \textbf{Normalization and Channel Recalibration}: Applies Batch Normalization (BN) for training stability, followed by a Squeeze-and-Excitation (SE) block \cite{hu2018squeeze} for adaptive channel-wise feature recalibration, optimizing feature utilization and model generalization.
    \item \textbf{Gate-Enhanced Feedforward Block (GEFB)}: Extends gated convolution techniques \cite{dauphin2017language}, improving feature extraction fidelity and processing efficiency. Its dynamic gating mechanism adaptively modulates feature flow, enhancing model robustness and scene-specific responsiveness.
\end{itemize}

This integrated backbone excels in feature extraction while exhibiting remarkable adaptability to complex scenarios and occlusions, significantly elevating our model's real-world performance.

\subsection{Multi-Scale Feature Integration and Upsampling}

The head layer transforms heterogeneous feature maps from various backbone blocks into a cohesive representation for precise pose estimation, employing fusion and progressive upsampling to synthesize multi-scale information.

Each backbone block $i$ produces a feature map $F_i \in \mathbb{R}^{C_i \times H_i \times W_i}$, where $C_i$, $H_i$, and $W_i$ denote channel depth and spatial dimensions. We employ the Dysample module \cite{liu2023learning} for efficient multi-scale feature fusion, resizing each feature map to a uniform target resolution $H_t \times W_t$:

\begin{equation}
F_i' = \text{Dysample}(F_i) \in \mathbb{R}^{C_i \times H_t \times W_t}
\end{equation}

The uniformly scaled feature maps are then fused along the channel axis:

\begin{equation}
F_{\text{fused}} = \text{Concat}(F_1', F_2', \ldots, F_n') \in \mathbb{R}^{C_{\text{total}} \times H_t \times W_t}
\end{equation}

where $C_{\text{total}} = \sum_{i=1}^{n} C_i$.

The aggregated $F_{\text{fused}}$ encapsulates a rich multi-scale representation, combining high-level semantics with fine-grained spatial details. This representation undergoes further refinement through convolutional layers with batch normalization and non-linear activations, enhancing its discriminative capacity.

The refined features are then fed into a task-specific prediction layer, translating the multi-scale, semantically-rich features into precise pose estimates resilient to occlusions and scene complexities.

This hierarchical fusion of multi-scale features enables effective synthesis of information across diverse spatial resolutions, yielding substantial improvements in pose estimation accuracy and robustness in challenging scenarios.

\subsection{Decoder and Loss Function}

The decoder transforms the fused feature representation into the final pose estimation heatmap through a cascade of operations:

\begin{equation}
K = \text{Conv}_{1 \times 1}(\text{Deconv}_2(\text{Deconv}_1(F_{\text{fused}})))
\end{equation}

where $F_{\text{fused}}$ is the concatenated multi-scale feature map, and $K \in \mathbb{R}^{J \times H \times W}$ represents the output heatmap for $J$ keypoints. Successive deconvolutions facilitate gradual upsampling, while the final $1 \times 1$ convolution distills the features into per-keypoint heatmaps.

Our loss function comprises output distillation loss and token-based knowledge distillation loss. The output distillation loss is:

\begin{equation}
\mathcal{L}_{\text{od}}^{t \rightarrow s} = \text{MSE}(K_s, K_t)
\end{equation}

where $K_s$ and $K_t$ are the heatmaps from student and teacher models, respectively.

The token-based knowledge distillation loss is defined as:

\begin{equation}
t^* = \arg \min_t \left(\text{MSE}(T(t; X), K_{\text{gt}})\right)
\end{equation}

where $T(t; X)$ represents transformed token predictions, and $K_{\text{gt}}$ is the ground truth heatmap.

These refinements enable our model to achieve state-of-the-art performance while maintaining efficiency.

\section{Experiments}

We evaluated GateAttentionPose on COCO \cite{lin2014microsoft} and MPII \cite{andriluka20142d} benchmarks, complemented by ablation studies to validate our design choices and elucidate component contributions.

\subsection{COCO Benchmark}

\begin{table}[!htbp]
    \centering
    \small
    \renewcommand{\arraystretch}{1.2}
    \resizebox{\columnwidth}{!}{
    \begin{tabular}{|l|c|c|c|ccc|ccc|}
        \hline
        \multirow{2}{*}{Method} & \multirow{2}{*}{Backbone} & \multirow{2}{*}{\makecell{Input\\size}} & \multirow{2}{*}{\makecell{Params\\(M)}} & \multicolumn{3}{c|}{COCO test-dev2017 $\uparrow$} & \multicolumn{3}{c|}{COCO val2017 $\uparrow$} \\
        \cline{5-10}
        & & & & AP & AP$^{50}$ & AP$^{75}$ & AP & AP$^{50}$ & AP$^{75}$ \\
        \hline
        SimBa. \cite{xiao2018simple} & ResNet-152 & 384$\times$288 & 68.6 & 73.7 & 91.9 & 81.1 & 74.3 & 89.6 & 81.1 \\
        PRTR \cite{li2021pose} & HRNet-W32 & 384$\times$288 & 28.5 & 71.7 & 90.6 & 79.6 & 73.1 & 89.4 & 79.8 \\
        TransPose \cite{yang2021transpose} & HRNet-W48 & 256$\times$192 & 17.5 & 75.0 & 92.2 & 82.3 & 75.8 & 90.1 & 82.1 \\
        TokenPose \cite{li2021tokenpose} & HRNet-W48 & 256$\times$192 & 63.6 & 75.9 & 92.3 & 83.4 & 75.8 & 90.3 & 82.5 \\
        HRNet \cite{sun2019deep} & HRNet-W48 & 384$\times$288 & 63.6 & 75.5 & 92.7 & 83.3 & 76.3 & 90.8 & 82.9 \\
        DARK \cite{zhang2020distribution} & HRNet-W48 & 384$\times$288 & 63.6 & 76.2 & 92.5 & 83.6 & 76.8 & 90.6 & 83.2 \\
        UDP \cite{huang2020devil} & HRNet-W48 & 384$\times$288 & 63.8 & 76.5 & 92.7 & 84.0 & 77.8 & 92.0 & 84.3 \\
        SimCC \cite{li20212d} & HRNet-W48 & 384$\times$288 & 66.0 & 76.0 & 92.4 & 83.5 & 76.9 & 90.9 & 83.2 \\
        HRFormer \cite{yuan2110hrformer} & HRFormer-B & 384$\times$288 & 50.3 & 76.2 & 92.7 & 83.8 & 77.2 & 91.0 & 83.6 \\
        ViTPose \cite{xu2022vitpose} & ViT-Base & 256$\times$192 & 85.8 & 75.1 & 92.5 & 83.1 & 75.8 & 90.7 & 83.2 \\
        SimBa. \cite{xiao2018simple} & Swin-Base & 256$\times$256 & 88.5 & 75.4 & 93.0 & 84.1 & 76.6 & 91.4 & 84.3 \\
        PCT \cite{geng2023human} & Swin-Base & 256$\times$192 & 220.1 & 76.2 & 92.1 & 84.5 & 77.2 & 91.2 & 84.3 \\
        BUCTD \cite{zhou2023rethinking} & HRNet-W48 & 256$\times$192 & 63.7 & 76.1 & 92.5 & 84.2 & 76.8 & 91.1 & 84.5 \\
        UniHCP \cite{ci2023unihcp} & ViT-Base & 256$\times$192 & 85.8 & 76.5 & 92.5 & 84.2 & 76.8 & 91.1 & 84.5 \\
        \hline
        Our approach & UniRepLKNet & 256$\times$192 & 61.1 & \textbf{76.9} & 90.7 & 83.4 & \textbf{77.4} & 90.7 & 84.2 \\
        \hline
    \end{tabular}
    }
    \caption{Performance comparison on COCO dataset}
    \label{tab:performance_comparison}
\end{table}

\textbf{Dataset:} The COCO dataset \cite{lin2014microsoft} comprises 57K training images (150K person instances), 5K validation images (6.3K person instances), and 20K test-dev images. We used the training set for model optimization and the validation set for performance assessment, reporting standard metrics including Average Precision (AP), AP\textsubscript{50}, and AP\textsubscript{75}.

\textbf{Results:} Table \ref{tab:performance_comparison} summarizes GateAttentionPose's performance against state-of-the-art methods. Our model achieves an AP of 76.9\% on COCO test-dev2017 and 77.4\% on COCO val2017, with a compact 61.1M parameters. This performance significantly outpaces other advanced approaches, demonstrating GateAttentionPose's efficacy in balancing accuracy and computational efficiency.

\subsection{MPII Benchmark}

\textbf{Dataset:} The MPII dataset \cite{andriluka20142d} contains ~25K images with over 40K annotated human poses. We used the standard train/test partition provided by the dataset curators, evaluating performance with PCKh (Percentage of Correct Keypoints with head-normalized distance).

\textbf{Results:} Table \ref{tab:keypoint_detection} compares GateAttentionPose against state-of-the-art methods on MPII. Our model exhibits superior PCKh across various joint categories, demonstrating robustness and generalizability across diverse pose configurations and environmental contexts.

\begin{table}[htbp]
    \begin{center}
    \renewcommand{\arraystretch}{1.5}
    \resizebox{\columnwidth}{!}{
    \begin{tabular}{|p{3cm}|c|c|c|c|c|c|c|c|}
        \hline
        Method & Hea. & Sho. & Elb. & Wri. & Hip. & Kne. & Ank. & Mean \\
        \hline
        SimBa. \cite{xiao2018simple} & 97.0 & 95.6 & 90.0 & 86.2 & 89.7 & 86.9 & 82.9 & 90.2 \\
        PRTR \cite{li2021pose} & 97.3 & 96.0 & 90.6 & 84.5 & 89.7 & 85.5 & 79.0 & 89.5 \\
        HRNet \cite{sun2019deep} & 97.1 & 95.9 & 90.3 & 86.4 & 89.7 & 88.3 & 83.3 & 90.3 \\
        DARK \cite{zhang2020distribution} & 97.2 & 95.9 & 91.2 & 86.7 & 89.7 & 86.7 & 84.7 & 90.3 \\
        TokenPose \cite{li2021tokenpose}  & 97.1 & 95.9 & 90.4 & 85.6 & 89.5 & 85.8 & 81.8 & 89.4 \\
        SimCC \cite{li20212d} & 97.2 & 96.0 & 90.4 & 85.6 & 89.5 & 85.8 & 81.8 & 90.0 \\
        \hline
        Our approach & 97.3 & 96.0 & 90.8 & 86.7 & 89.4 & 86.6 & 82.3 & \textbf{90.6} \\
        \hline
    \end{tabular}
    }
    \end{center}
    \caption{Performance comparison on MPII dataset}
    \label{tab:keypoint_detection}
\end{table}

\subsection{Ablation Studies}

We conducted comprehensive ablation studies on the COCO dataset to quantify the impact of four key components in GateAttentionPose: GLACE module optimization, Agent Attention mechanism, Gate-Enhanced Feedforward Block (GEFB), and Dysample operation. Average Precision (AP) served as the primary evaluation metric.

\begin{table}[htbp]
    \begin{center}
    \resizebox{\columnwidth}{!}{%
    \begin{tabular}{|c|c|c|c|c|c|}
        \hline
        UniRepLKNet & GLACE & Agent Attention & GEFB & Dysample & COCO/AP \\
        \hline
        $\checkmark$ & & & & & 75.3 \\
        \hline
        & $\checkmark$ & & & & 75.5 \\
        \hline
        & & $\checkmark$ & & & 75.4 \\
        \hline
        & & & $\checkmark$ & & 76.0 \\
        \hline
        & $\checkmark$ & $\checkmark$ & & & 75.8 \\
        \hline
        & $\checkmark$ & & $\checkmark$ & & 76.4 \\
        \hline
        & $\checkmark$ & $\checkmark$ & $\checkmark$ & & 76.8 \\
        \hline
        & $\checkmark$ & $\checkmark$ & $\checkmark$ & $\checkmark$ & 76.9 \\
        \hline
    \end{tabular}
    }
    \end{center}
    \caption{Ablation Study on COCO Dataset}
    \label{tab:ablation_study}
\end{table}
\textbf{Analysis and Results:} Our ablation studies, starting from a baseline UniRepLKNet (AP 75.3), reveal:

\begin{itemize}
    \item GLACE module optimization significantly improves feature map embedding.
    \item Agent Attention enhances computational efficiency while maintaining global context modeling.
    \item Gate-Enhanced Feedforward Block (GEFB) refines feature extraction, especially in complex scenarios.
    \item Dysample upsampling technique contributes to overall performance.
\end{itemize}

The integration of all components results in a peak AP of 76.9. Table \ref{tab:ablation_study} summarizes these findings, demonstrating each element's contribution to GateAttentionPose's improved accuracy.

\subsection{Results Analysis}

GateAttentionPose demonstrates superior efficacy in human pose estimation on COCO:

\begin{itemize}
    \item Synergistic integration of GLACE module, Agent Attention, and GEFB enables state-of-the-art accuracy with computational efficiency.
    \item The model achieves a favorable balance between precision and resource utilization, beneficial for resource-constrained environments.
    \item High AP scores validate the model's capability in capturing fine-grained pose information across diverse scenarios.
    \item Results indicate strong potential for real-world applications with environmental variability and computational constraints.
\end{itemize}

These outcomes underscore GateAttentionPose's robust performance and adaptability for both experimental and practical deployment in human pose estimation tasks.

\section{Conclusion}

We introduce GateAttentionPose, an innovative pose estimation framework combining Agent Attention and advanced gated convolution techniques. Key contributions include:

\begin{itemize}
    \item Integration of Agent Attention for enhanced computational efficiency and global context modeling.
    \item Introduction of Gate-Enhanced Feedforward Block (GEFB) for improved feature extraction.
    \item State-of-the-art performance on COCO and MPII datasets with a compact model architecture.
    \item Superior accuracy and reduced computational footprint compared to contemporary methods.
\end{itemize}

GateAttentionPose effectively handles complex scenes with occlusions and variable illumination, balancing accuracy and efficiency. This makes it suitable for resource-constrained real-world applications.

Our work advances pose estimation and lays groundwork for future computer vision innovations, potentially inspiring further optimizations in visual understanding tasks.

\section*{Acknowledgment}

The author is deeply grateful to Prof. Ming Xu for his invaluable guidance and support throughout this research.

\bibliographystyle{IEEEtran}
\bibliography{references}

\end{document}